\documentclass{article}
\usepackage{tikz}
\usepackage{xcolor}
\usepackage{amssymb}

\definecolor{lightblue}{RGB}{173,216,230}

\DeclareRobustCommand{\blueTri}{%
  \mathbin{%
    \tikz[baseline=-0.5ex, scale=0.2]{%
      \fill[lightblue]
        (0,0) -- (0.9,0) -- (0.45,0.78) -- cycle;
      \draw[gray, line width=0.5pt]
        (0,0) -- (0.9,0) -- (0.45,0.78) -- cycle;
    }%
  }%
}

\usepackage{hyperref}

\usepackage{microtype}
\usepackage{graphicx}
\usepackage{subcaption}
\usepackage{booktabs}
\usepackage{multirow}
\usepackage{float}
\usepackage{caption}
\usepackage{wrapfig}
\usepackage[table]{xcolor}
\usepackage[most]{tcolorbox}
\definecolor{darkblue}{RGB}{24,112,176}
\usepackage[export]{adjustbox}
\usepackage{bbding}
\newtcolorbox{promptbox}[2][]{%
  enhanced,
  breakable,
  colback=gray!6,
  colframe=darkblue!80,
  boxrule=0.6pt,
  arc=2pt,
  left=6pt,right=6pt,top=6pt,bottom=6pt,
  fonttitle=\bfseries,
  title={#2},
  #1
}

\definecolor{codegreen}{rgb}{0,0.5,0}
\definecolor{codegray}{rgb}{0.5,0.5,0.5}
\definecolor{codepurple}{rgb}{0.58,0,0.82}
\definecolor{backcolour}{rgb}{1.0,1.0,1.0}
\definecolor{link}{rgb}{1.0,1.0,1.0}
\lstdefinestyle{mystyle}{
    backgroundcolor=\color{backcolour},   
    commentstyle=\color{codegreen},
    keywordstyle=\color{magenta},
    numberstyle=\tiny\color{codegray},
    stringstyle=\color{codepurple},
    basicstyle=\ttfamily\footnotesize,
    breakatwhitespace=false,         
    breaklines=true,                 
    keepspaces=true,                 
    numbers=left,                    
    numbersep=5pt,                  
    showspaces=false,                
    showstringspaces=false,
    showtabs=false,                  
    tabsize=1,
    float=tp,
  floatplacement=tbp
}
\DeclareCaptionFormat{mylst}{\hrule#1#2#3\hrule}
\usepackage[main,final]{neurips_2026}

\usepackage{amsmath}
\usepackage{amssymb}
\usepackage{mathtools}
\usepackage{amsthm}
\usepackage{multirow}
\usepackage{multicol}
\usepackage{colortbl}

\newcommand*{\myalign}[2]{\multicolumn{1}{#1}{#2}}

\usepackage{booktabs}
\usepackage{enumitem}

\definecolor{blue}{HTML}{1F77B4} 
\definecolor{green}{HTML}{2CA02C} 
\definecolor{teal}{HTML}{17BECF} 
\definecolor{lightblue}{HTML}{b0d0e0}

\usepackage[capitalize,noabbrev]{cleveref}

\theoremstyle{plain}

\theoremstyle{definition}

\theoremstyle{remark}

\usepackage[textsize=tiny]{todonotes}

\title{Test-Time Scaling with Diffusion Language Models via \\ Reward-Guided Stitching}

\author{
Roy Miles \\
MVP Lab \\
\texttt{} \\
\And
Aysim Toker \\
MVP Lab \\
\texttt{} \\
\And
Andreea-Maria Oncescu \\
MVP Lab \\
\texttt{} \\
\And
Jiankang Deng\NoHyper\thanks{Corresponding author: jiankangdeng@gmail.com}\endNoHyper \\
MVP Lab \\
\texttt{} \\
\And
Ismail Elezi \\
MVP Lab \\
\texttt{} \\
}

\begin{document}

\maketitle




\begin{abstract}
Reasoning with large language models often benefits from generating multiple chains-of-thought, but existing aggregation strategies are typically trajectory-level (e.g., selecting the best trace or voting on the final answer), discarding useful intermediate work from partial or “nearly correct” attempts. We propose Stitching Noisy Diffusion Thoughts, a self-consistency framework that turns cheap diffusion-sampled reasoning into a reusable pool of step-level candidates. Given a problem, we (i) sample many diverse, low-cost reasoning trajectories using a masked diffusion language model, (ii) score every intermediate step with an off-the-shelf process reward model (PRM), and (iii) stitch these highest-quality steps across trajectories into a composite rationale. This rationale is then used to recompute only the final answer. This modular pipeline separates exploration (diffusion) from evaluation and solution synthesis, avoiding monolithic unified hybrids while preserving broad search. Across math reasoning benchmarks, we find that step-level recombination is most beneficial on harder problems, and ablations highlight the importance of the final solver in converting stitched but imperfect rationales into accurate answers. Using low-confidence diffusion sampling with parallel, independent rollouts, our training-free framework improves average accuracy by up to 23.8\% across six math and coding tasks. At the same time, it achieves up to a 1.8× latency reduction relative to both traditional diffusion models (e.g., Dream, LLaDA) and unified architectures (e.g., TiDAR). The code is available \href{https://github.com/roymiles/diffusion-stitching}{here}.

\end{abstract}

\section{Introduction}
Large language models have become strong general-purpose reasoners, but reliable reasoning at test time is still expensive. On hard problems, accuracy often scales with the amount of computation devoted to search and naive scaling quickly becomes impractical. A core difficulty is that reasoning traces are noisy: a single incorrect intermediate step can derail an otherwise promising derivation, so allocating all compute to one long trajectory is risky. This motivates approaches that trade a single high-fidelity rationale for a broader exploration under a fixed latency budget.

More broadly, improving the practicality of large models requires addressing resource costs across both training and inference; complementary work has reduced the memory overhead of LLM training~\cite{miles2024veloramemoryefficienttraining, hu2021loralowrankadaptationlarge}, while we focus here on inference-time latency.

\begin{figure*}
    \centering
    \includegraphics[width=.95\linewidth,keepaspectratio]{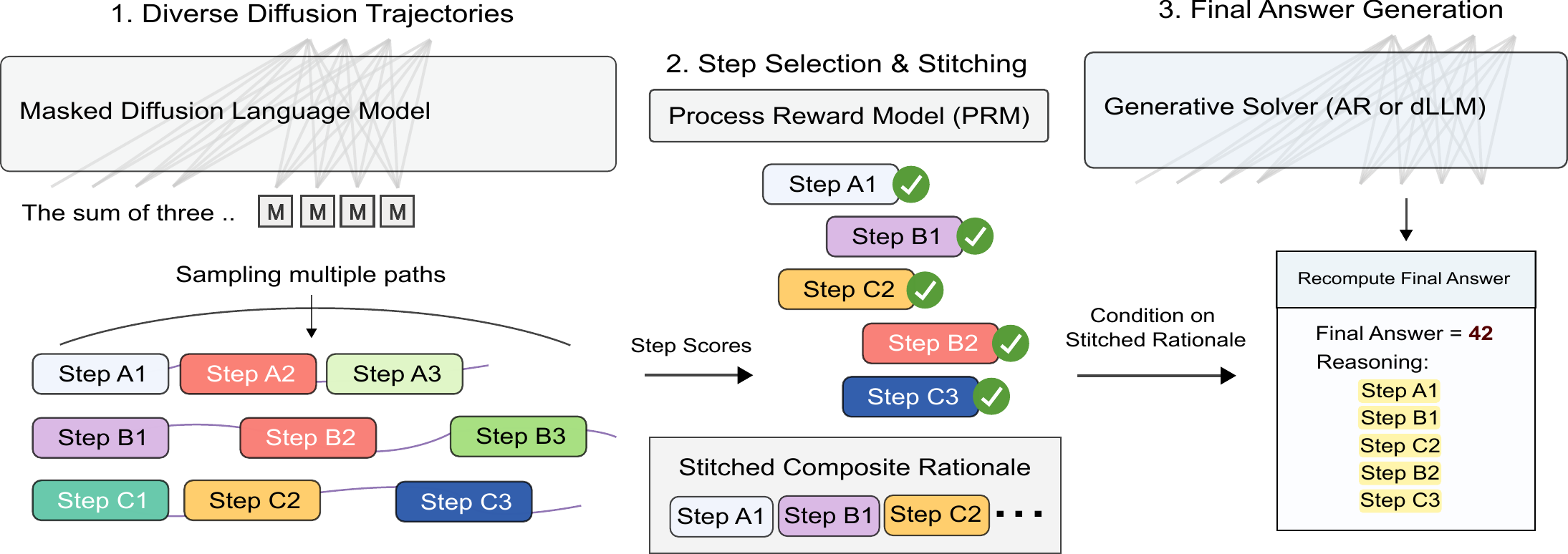}
    \caption{\textbf{Diffusion Stitching pipeline}. We first use a diffusion model to efficiently explore diverse reasoning paths, then score each intermediate step with a PRM, and finally stitch the highest-quality steps into a single rationale that conditions a generative solver to recompute the final answer.}
    \label{fig:pipeline}
\end{figure*}

A common strategy is to sample multiple reasoning paths and aggregate them, e.g., via self-consistency or verifier-based selection~\cite{zhang2025generative}. However, most aggregation methods operate at the trajectory level: they choose one full chain (or vote over final answers), effectively discarding partial work from runs that contain useful intermediate results but fail later. In parallel, diffusion language models offer an appealing mechanism for broad exploration because they can refine partially masked sequences with parallel denoising, producing many diverse, low-cost reasoning trajectories. 
Recent unified hybrid models~\cite{Liu2025TiDAR} that combine diffusion-style refinement with autoregressive (AR) decoding can be effective, but they couple the exploration, scoring, and final generation stages together in a way that limits flexibility and makes targeted improvements harder. In practice, they also see a big drop in reasoning accuracy compared to the strong AR models.

In this work, we propose Stitching Noisy Diffusion Thoughts, a self-consistency framework that turns diffusion sampling into a reusable pool of intermediate reasoning steps rather than a set of complete candidate solutions. Our key observation is that cheap diffusion-sampled chains can contain many locally correct sub-results. If we can identify these high-utility steps across trajectories and recombine them, we can recover much of the benefit of a broad search without paying for a single long derivation from scratch. Concretely, we advocate a specialist pipeline: a diffusion model for inexpensive exploration, a step-level evaluator to score intermediate reasoning, and a lightweight autoregressive model to produce the final answer conditioned on the stitched rationale.

Our method has three components. (i) Explore: given an input problem, we sample $N$ diverse chains-of-thought using a masked diffusion language model with a confidence-based sampling procedure that encourages diversity while preserving high-confidence content. (ii) Evaluate: we score every intermediate step in every trajectory using an off-the-shelf process reward model (PRM), yielding a global pool of candidate steps with quality estimates. This step-level view is crucial: it allows us to retain useful intermediate results even when an overall trajectory later derails. (iii) Stitch + Recompute: rather than selecting a single “best” trajectory (e.g., by maximizing the geometric mean of step scores), we collect the highest-quality steps across paths and concatenate them into a composite rationale. A small autoregressive model then conditions on the original problem and this stitched rationale and recomputes the final answer. 

Empirically, this separation of roles makes the system both modular and effective. The diffusion component supplies breadth; the PRM provides a fine-grained notion of step quality; and the AR solver converts a partially redundant set of high-quality steps into a more accurate final response. This design also cleanly exposes ablations: we can quantify the value of (a) diffusion exploration alone, (b) trajectory-level selection, and (c) step-level stitching plus final answer generation. In summary, our contributions are given as follows:
\begin{enumerate}
    \item We \textbf{introduce} a step-level framework for reasoning that replaces trajectory-level selection with PRM-scored step selection and stitching over diffusion-sampled chains-of-thought. 
    \item We \textbf{analyze} key stitching choices: generation count, diffusion diversity (e.g., temperature), PRM thresholding/anchoring, and the model choice for the generative solver. In doing so we show how each component improves robustness and test-time scaling.
    \item We \textbf{demonstrate} strong reasoning under tight latency budgets, improving the accuracy--latency Pareto frontier over state-of-the-art AR and diffusion baselines (see Figure \ref{fig:teaser} and \ref{fig:pareto}), with up to \textbf{9.85$\times$} fewer sequential forward passes / \textbf{1.8$\times$} lower end-to-end latency at matched accuracy, and up to \textbf{+30.6\%} absolute accuracy points over vanilla diffusion decoding.
\end{enumerate}

\vspace{-0.8em}
\section{Related Work}
\paragraph{Diffusion Language Models.}
Masked diffusion language models (dLLMs)~\cite{Nie2025LLaDA} formulate generation as iterative denoising of partially masked sequences, leveraging bidirectional Transformers for parallel refinement~\cite{Devlin2019BERT,Austin2021D3PM,Hoogeboom2021Multinomial,Ghazvininejad2019MaskPredict}.
In text generation, diffusion has been explored for controllable and conditional settings~\cite{Li2022DiffusionLM,Gong2022DiffuSeq,lou2024discretediffusionmodelingestimating}, and recent scaling efforts show competitiveness with autoregressive (AR) models, e.g., LLaDA and Dream~\cite{Nie2025LLaDA,Ye2025Dream}.
On the inference side, Fast-dLLM~\cite{wu2025fastdllmtrainingfreeaccelerationdiffusion} accelerates decoding via approximate KV caching and parallelization.
For reasoning, diffusion enables diverse chain-of-thought (CoT) sampling aggregated via self-consistency~\cite{Ye2024DoT,wang2023selfconsistencyimproveschainthought,Shao2025EfficientThought}.
Recent work also explores scaling (LLaDA 2.0)~\cite{bie2025llada20scalingdiffusionlanguage} and hybrid architectures such as Block Diffusion and TiDAR~\cite{arriola2025blockdiffusioninterpolatingautoregressive,Liu2025TiDAR}.
In contrast, we advocate a decoupled pipeline where diffusion enables exploration and a lightweight AR model handles the final answer generation.

\vspace{-10pt}
\paragraph{Speculative Decoding.}
Speculative decoding accelerates AR generation via draft-and-verify: a lightweight model proposes tokens, and a larger model verifies them in parallel~\cite{leviathan2022speculative,chen2023speculative}.
Recent methods replace the AR drafter with diffusion models to reduce sequential bottlenecks, including SpecDiff, SpecDiff-2, DiffuSpec, and DEER~\cite{christopher-etal-2025-speculative,sandler2025specdiff2scalingdiffusiondrafter,li2025diffuspecunlockingdiffusionlanguage,cheng2025deerdraftdiffusionverify}.
Autospeculation approaches such as Spiffy and SSD further allow diffusion models to draft and verify their own predictions~\cite{agrawal2026spiffymultiplyingdiffusionllm,gao2025selfspeculativedecodingdiffusion}.
Overall, these methods focus on token-level alignment via acceptance mechanisms, whereas we target reasoning by exploring and recombining trajectory-level intermediate steps.

\vspace{-10pt}
\paragraph{Test-Time Reasoning.}
Test-time compute improves reasoning in LLMs through methods such as chain-of-thought prompting~\cite{wei2022chainofthought}, self-consistency~\cite{wang2023selfconsistencyimproveschainthought}, least-to-most prompting~\cite{zhou2022leasts}, and Tree-of-Thought~\cite{yao2023treethoughtsdeliberateproblem}.
Verifier-based approaches further select high-quality candidates~\cite{cobbe2021trainingverifiers}, while interactive methods like ReAct introduce iterative reasoning with feedback~\cite{yao2023react}.
However, these approaches often incur significant computational overhead~\cite{feng2025optimalselfconsistencyefficientreasoning}.
We instead leverage diffusion to generate diverse reasoning trajectories in parallel and stitch high-quality steps, enabling broad exploration with reduced latency.
\vspace{-0.5em}
\section{Method}
\vspace{-0.5em}
Improved reasoning often scales with heavy test-time compute, making it impractical for low-latency applications. This issue stems from treating each reasoning trace as an all-or-nothing object, which both wastes useful partial progress and makes aggregation high-variance. 
In this section, we propose a framework that addresses this problem by reframing reasoning as a modular pipeline using both efficient search and robust step evaluation (see~\cref{fig:pipeline}). Our approach first decouples exploration from synthesis: we first generate a diverse pool of low-cost thoughts via a diffusion model (\cref{sec:method_exploration}), evaluate their utility at the step level using a process reward model (\cref{sec:method_scoring}), and finally employ a stitching mechanism (\cref{sec:method_stitching}) to recombine high-confidence steps into a coherent rationale that guides the generative solver to recompute the final answer (\cref{sec:method_recompute}).

\subsection{Preliminaries: Masked Diffusion Language Models.}
LLaDA~\cite{Nie2025LLaDA} produces an $L$-token sequence via iterative refinement, progressively replacing mask tokens in an initially fully-masked template with predicted tokens, rather than generating tokens one-by-one left-to-right as in autoregressive models. Let $\mathcal{V}$ be the vocabulary, $M$ denote the mask token, and initialize the input sequence $x^{(0)}=[\text{prompt}, M,\dots,M]\in(\mathcal{V}\cup\{M\})^{L}$. For diffusion steps $k=1,\dots,K$, the bidirectional mask predictor outputs logits $\ell^{(k)}_{i}\in\mathbb{R}^{|\mathcal{V}|}$ for each masked position $i$, which we convert to a sampling distribution with \emph{temperature} $\tau>0$:
\begin{align}
    p^{(k)}_{i} =\mathrm{softmax}\!\left(\ell^{(k)}_{i}/\tau\right),\qquad \hat{x}^{(k)}_i \sim p^{(k)}_{i}.
\end{align}
We define a per-position \emph{confidence} score as $c^{(k)}_i=\max_{j} p^{(k)}_{i,j}$. LLaDA then performs \emph{low-confidence remasking}: after proposing $\hat{x}^{(k)}_i$ for masked positions, it keeps high-confidence predictions fixed while remasking uncertain ones, e.g. remask all positions with $c^{(k)}_i<\gamma$ (or equivalently, remask the lowest-confidence tokens to match a target mask budget), where $\gamma\in[0,1]$ is a tunable \emph{confidence} threshold. Repeating this refine--(re)mask process until no masks remain yields the final completion. This iterative process is unlike autoregressive decoding, which commits to a left-to-right prefix. 

\begin{figure*}[!t]
    \centering
    \includegraphics[width=.7\linewidth]{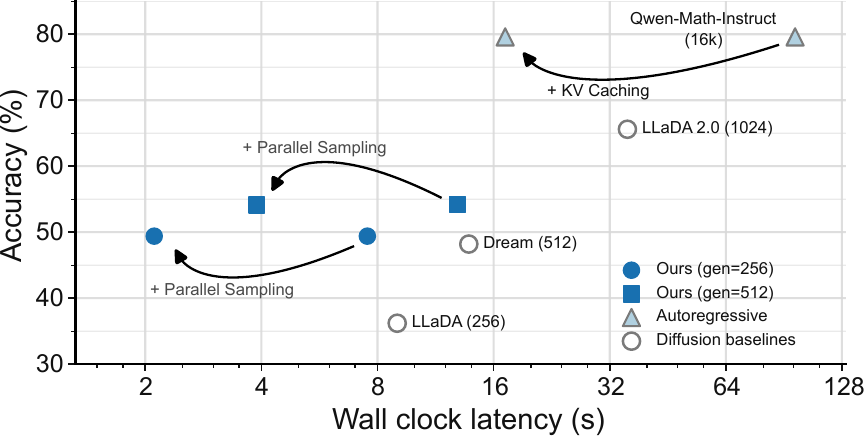}
    \caption{Accuracy vs. wall-clock latency on \textsc{Math500} comparing autoregressive baselines ($\blueTri$) using early stopping, diffusion baselines ($\color{gray} \circ$), and our stitching pipeline ($\color{blue} \blacksquare$, $\color{blue} \bullet$) using 4 reasoning traces. All models are 7B in size, except for LLaDA 2.0. All diffusion models are run without KV caching, so enabling caching~\cite{wu2025fastdllmtrainingfreeaccelerationdiffusion} would likely provide additional practical speedups.}
    \label{fig:teaser}
\end{figure*}

\subsection{Reasoning Path Exploration.}
\label{sec:method_exploration}
Masked diffusion language models are well suited to broad exploration: starting from a partially-masked sequence, they iteratively denoise tokens in parallel, rather than generating tokens in a strict chronological order.
Given an input problem $x$, we run $N$ independent diffusion generations to obtain diverse reasoning traces. Each trace $n\in\{1,\dots,N\}$ is then deterministically segmented into human-readable steps:
\begin{align}
    \tau^{(n)} \;=\; \bigl(s^{(n)}_1,\dots,s^{(n)}_{T_n}\bigr),
    \qquad
    s^{(n)}_t \in \mathcal{V}^*.
\end{align}

We segment each generated rationale into a sequence of steps using task-appropriate boundaries. For example, for the math problems, we split into sentence-level steps.

\subsection{Evaluating the quality of each reasoning step.}
\label{sec:method_scoring}
Given the segmented reasoning traces $\{\tau^{(n)}\}_{n=1}^N$, we assign a quality score to every step using an off-the-shelf process reward model (PRM)~\cite{prmlessons,zeng2025acecoderacingcoderrl}.
For each trace $n$ and step index $t$, the PRM outputs a scalar confidence:
\begin{align}
    r^{(n)}_t \;=\; \mathrm{PRM}_\phi\!\bigl(x,\; s^{(n)}_{1:t}\bigr)\in[0,1],
\end{align}
where the score is conditioned on the problem $x$ and the partial solution history $s^{(n)}_{1:t}$.
This step-level view allows us to retain useful intermediate derivations even when a reasoning path later derails.

\paragraph{Single-pass scoring via marker tokens.}
In practice, we score all steps in a trace using one forward pass by inserting a special marker token $\langle m\rangle$
(e.g., \texttt{<extra\_0>}) after each step boundary and reading the model's predicted probability of the marker:
\begin{align}
    r^{(n)}_t \;=\; p_\phi\!\left(\langle m\rangle \,\middle|\, x,\; s^{(n)}_{1:t}\right).
\end{align}
Collecting scores over all traces yields a global pool of scored steps,
\begin{align}
    \label{eqn:scored_step_pool}
    \mathcal{P} \;=\; \bigl\{(n,t, s^{(n)}_t, r^{(n)}_t)\;:\; n\in[N],\; t\in[T_n]\bigr\},
\end{align}
which forms the candidate set for the stitching process in~\cref{sec:method_stitching}.

\subsection{Stitching the reasoning steps together.}
\label{sec:method_stitching}
\begin{figure}
    \centering
    \includegraphics[width=\linewidth]{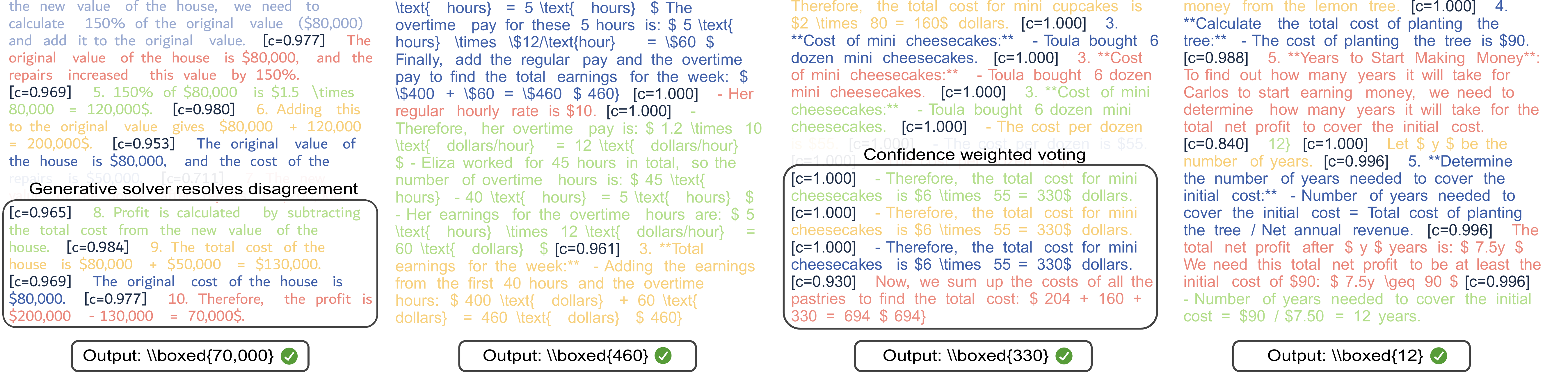}
    \caption{  \textbf{Qualitative examples of step stitching}. Each column shows a stitched evidence trace built from multiple diffusion trajectories (colors denote retained high-scoring steps). A generative solver then conditions on this trace to output only the final answer. Column 1 illustrates resolving disagreements; column 3 shows confidence scores guide the final prediction.}
    \label{fig:qualitative}
\end{figure}

From the global pool of scored steps in~\cref{eqn:scored_step_pool},
we build a stitched \emph{evidence list} $E$ by selecting reliable steps across all reasoning traces.
We score each trace by the geometric mean of its step-level PRM scores. Let $n^\star$ denote the trace with the highest geometric-mean score. We always keep the \emph{final step} of this best trace as an answer anchor.
Given a confidence threshold $\delta \in[0,1]$, we retain all steps in $\mathcal{P}$ with a PRM score at least $\delta$, and we also include every step from the best full trace (to serve as an anchor).
We construct the stitched rationale as a sequence of $(\text{step}, \text{confidence})$ pairs, i.e., by prefixing each retained step with its score (\texttt{[c=0.93] ...}), which allows the downstream solver to condition jointly on the content and its confidence score (see Figure \ref{fig:qualitative}). To maintain coherence, we then concatenate selected steps in chronological order based on their original positions. This preserves local dependencies while still allowing complementary sub-derivations from different reasoning traces to co-exist in a single rationale. Finally, we utilize the resulting confidence-annotated evidence list $E$ to recompute the final answer (\cref{sec:method_recompute}).

\subsection{Recomputing the final answer.}
\label{sec:method_recompute}
The stitched evidence list $E$ is not guaranteed to form a complete and perfectly consistent chain-of-thought:
it may contain redundancy, small gaps, or occasional contradictions from mixing reasoning paths.
Instead of directly returning the endpoint of any diffusion trace, we use a lightweight autoregressive (AR) solver to \emph{recompute} the final answer.

\paragraph{Evidence-conditioned decoding.}
We construct a solver prompt by concatenating the problem $x$ with the confidence-annotated evidence list $E$. We explicitly instruct the solver to treat these steps as evidence, prioritizing high-confidence entries while ignoring conflicts. We then obtain the final answer $\hat{y}$ by maximizing the likelihood:
\[
\hat{y} \;=\; \arg\max_{y} p_{\psi}(y \mid x, E),
\]
implemented via greedy or low-temperature decoding with a strict stopping rule (e.g., stop after producing a complete \texttt{\textbackslash boxed\{...\}} answer) with $\psi$ denoting the parameters of the generative solver.
In effect, this AR stage acts as a reconciliation step: it selects a consistent subset of evidence, fills in missing links, and produces a coherent final solution.
In practice, we implement this via greedy or low-temperature decoding with a strict stopping rule.

\subsection{Extending to coding tasks.}
\begin{figure}
    \centering
    \includegraphics[width=.7\linewidth]{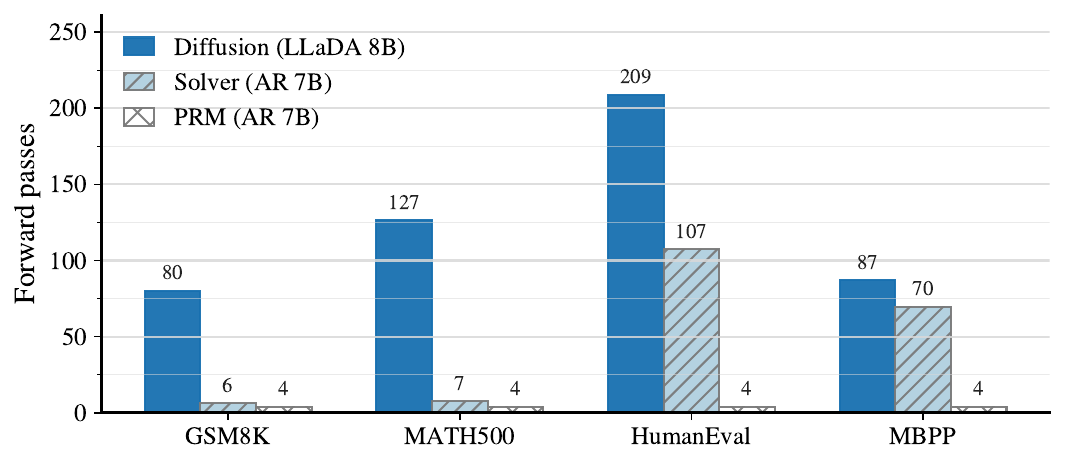}
    \caption{\textbf{Parallisable inference cost (gen length 512)}. We report the number of parallelisable diffusion steps and the number of AR solver decoding steps for math and coding tasks. The results covering only diffusion generation (reasoning and answer) are provided in table \ref{tab:diffusion_solver}.}
    \label{fig:forward_passes_bar}
\end{figure}

The same pipeline extends to code generation by redefining steps and using code-aware evaluators. We use different step definitions for MBPP and HumanEval: MBPP solutions are short and benefit from local code-line reuse, whereas HumanEval often requires more global planning, for which natural-language rationales are more robust.
For MBPP, we treat each line in the code as a step. The diffusion model then proposes candidate code, a code PRM assigns a confidence score to each line, and we stitch high-confidence lines together. An autoregressive solver then generates the final program conditioned on this stitched evidence, correcting inconsistencies and filling in missing details. As shown in~\cref{fig:forward_passes_bar}, this conditioning yields concise solutions with 3.21× fewer forward passes than the baseline.
For HumanEval, we instead stitch a natural-language rationale and condition the solver on this reasoning before generating code, which avoids spending compute on re-generating the rationale and allows the solver to just generate the code implementation.
\section{Experiments}
\begin{table*}[ht]
\centering
\scriptsize
\caption{\textbf{Evaluation Results:} We evaluate the performance of Diffusion Stitching over several coding and math tasks. Baseline LLaDA and Dream numbers are taken from the TiDAR paper. The confidence-sampling variants were reproduced using $\gamma = 0.9$ and 512 denoising steps, respectively. For our pipeline$^\star$ the 8B size refers to the diffusion backbone and all runs were evaluated in a zero-shot setting.}
\begin{tabular}{llcccc|cc|c}
\toprule
 & & \multicolumn{4}{c}{\textbf{Coding}} & \multicolumn{2}{c}{\textbf{Math}} & \multicolumn{1}{c}{\textbf{Avg}} \\
\midrule
\textbf{Model Arch} & \textbf{Size} & \multicolumn{1}{c}{\textbf{HumanEval}} & \multicolumn{1}{c}{\textbf{HumanEval+}} & \multicolumn{1}{c}{\textbf{MBPP}} & \multicolumn{1}{c}{\textbf{MBPP+}} & \multicolumn{1}{c}{\textbf{GSM8K}} & \multicolumn{1}{c}{\textbf{Minerva Math}} & \multicolumn{1}{c}{} \\
\midrule
Qwen3 & 4B & 57.32 & 50.61 & 67.00 & 80.69 & 77.48 & 47.10 & 63.37 \\
Qwen3 & 8B & 64.63 & 56.71 & 69.40 & 83.07 & 81.80 & 52.94 & 68.09 \\
\midrule
LLaDA & 8B & 32.32 & 27.44 & 40.80 & 51.85 & 70.96 & 27.30 & 41.78 \\
\myalign{l}{\;\;\;\footnotesize $\rotatebox[origin=c]{180}{$\Lsh$}$ w/ conf sampling} & 8B & 40.24 & 37.80 & 41.89 & 55.82 & 78.77 & 36.54 & 48.51 \\
Dream & 7B & 54.88 & 49.39 & 56.80 & 74.60 & 77.18 & 39.60 & 58.74 \\
{\;\;\;\footnotesize $\rotatebox[origin=c]{180}{$\Lsh$}$ w/ conf sampling} & 7B & 59.76 & 57.32 & 59.14 & 71.96 & 81.96 & 42.78 & 62.15 \\
Block Diff & 4B$^{\dagger}$ & 56.10 & 51.22 & 54.60 & 69.84 & 82.87 & 47.02 & 60.27 \\
TiDAR (Trust AR) & 8B$^{\ddagger}$ & 55.49 & 52.44 & 65.40 & 79.63 & 79.83 & 50.58 & 63.90 \\
TiDAR (Trust Diff) & 8B$^{\ddagger}$ & 57.93 & 55.49 & 65.40 & 80.95 & 80.44 & 51.64 & 65.31 \\
\rowcolor{cyan!15} Ours$^\star$ & 8B & \textbf{70.37} & \textbf{64.02} & \textbf{74.61} & \textbf{81.75} & \textbf{91.51} & \textbf{53.20} & \textbf{72.38} \\
\midrule
\end{tabular}
\label{tab:tidar}
\end{table*}
\begin{figure*}[ht]
    \centering
    \begin{subfigure}{0.33\textwidth}
        \centering
         \includegraphics[width=\linewidth]{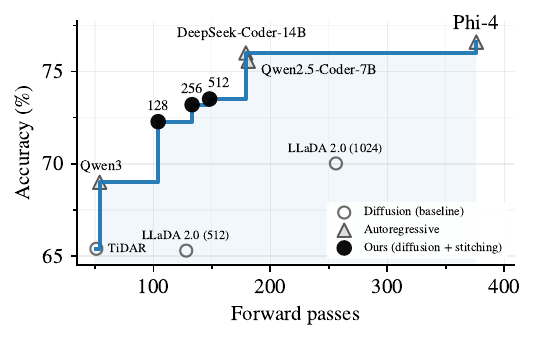}
        \subcaption{MBPP}
        \label{fig:stitching-mbpp}
    \end{subfigure}\hfill
    \begin{subfigure}{0.33\textwidth}
        \centering
        \includegraphics[width=\linewidth]{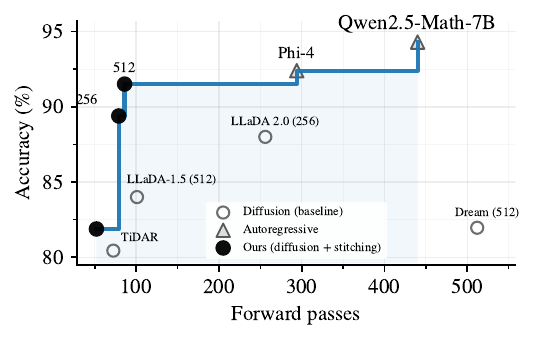}
        \subcaption{GSM8K}
        \label{fig:stitching-gsm8k}
    \end{subfigure}\hfill
    \begin{subfigure}{0.33\textwidth}
        \centering
        \includegraphics[width=\linewidth]{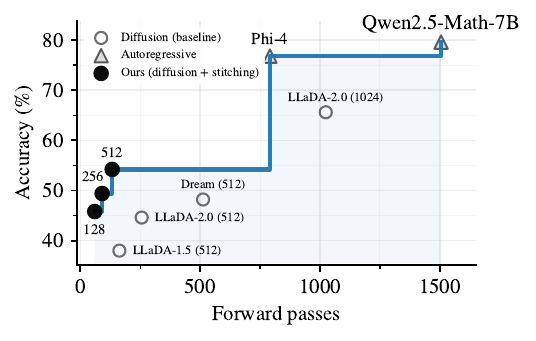}
        \subcaption{MATH500}
        \label{fig:stitching-math500}
    \end{subfigure}

    \caption{\textbf{Pareto fronts of forward passes vs. accuracy} on MBPP, GSM8K, and MATH500, showing that diffusion stitching bridges the gap between diffusion and AR models while enabling efficient test-time scaling for low-latency reasoning. All models are of the same 7-8B size, except for DeepSeek Coder which is 14B. We report our result for generation length 128, 256, and 512.}
    \label{fig:pareto}
\end{figure*}

\paragraph{Implementation details.}
We distribute the batch of $N$ independent diffusion generations over $G$ GPUs, assigning [$N/G$] traces per device and accumulating step-level PRM scores locally. 
All latency experiments were conducted using up to four NVIDIA H200 GPUs with PyTorch in bfloat16; independent diffusion rollouts were distributed across GPUs.
Following this local evaluation, we perform a single \texttt{all\_gather} operation to synchronize the scores and content, thereby forming the global pool of candidate steps. Since the trajectories are generated independently, our method scales linearly with the available compute, requiring synchronization only for the final aggregation.

\begin{wraptable}{r}{0.55\columnwidth}
\centering
\scriptsize
\caption{\textbf{Accuracy--latency comparison against AR baselines.}
We compare against Qwen2.5-Math-7B-Instruct using greedy decoding,
self-consistency, and PRM best-of-4 under the same hardware setup.}
\vspace{0.3em}
\label{tab:ar_latency}

\setlength{\tabcolsep}{2.2pt}
\renewcommand{\arraystretch}{0.92}

\resizebox{\linewidth}{!}{%
\begin{tabular}{lcccccc}
\toprule
& &
\multicolumn{2}{c}{\textbf{GSM8K}} &
\multicolumn{2}{c}{\textbf{MATH500}} &
\\
\cmidrule(lr){3-4}
\cmidrule(lr){5-6}

\textbf{Setting} &
\textbf{Len.} &
\textbf{Acc.} &
\textbf{Lat.} &
\textbf{Acc.} &
\textbf{Lat.} &
\textbf{Mem.} \\

& &
(\%) &
(s) &
(\%) &
(s) &
(GB) \\
\midrule

\rowcolor{cyan!15}
Ours (Qwen PRM)
& --
& 91.3
& 2.585
& 54.3
& 4.735
& 42.8 / 42.7 \\

Ours (Skywork PRM)
& --
& 89.4
& 2.421
& 51.8
& 3.910
& 32.3 / 32.2 \\

\midrule
\rowcolor{gray!12}
\multicolumn{7}{l}{\emph{Greedy}} \\

Qwen2.5-Math-7B
& 256
& 32.0
& 2.545
& 18.0
& 4.989
& 14.4 / 14.4 \\

&
512
& 84.0
& 3.250
& 49.4
& 8.869
& 14.4 / 14.4 \\

&
1024 / 768
& 90.0
& 3.829
& 64.2
& 9.849
& 14.5 / 14.5 \\

\midrule
\rowcolor{gray!12}
\multicolumn{7}{l}{\emph{Self-consistency ($N=4$)}} \\

Qwen2.5-Math-7B
& 256
& 34.0
& 2.798
& 20.0
& 5.310
& 14.4 / 14.4 \\

&
512
& 90.0
& 3.932
& 56.4
& 9.210
& 14.4 / 14.5 \\

&
1024 / 768
& 92.0
& 4.361
& 69.6
& 10.130
& 14.5 / 14.5 \\

\midrule
\rowcolor{gray!12}
\multicolumn{7}{l}{\emph{PRM best-of-4}} \\

Qwen2.5-Math-7B
& 256
& 34.0
& 2.990
& 18.0
& 5.311
& 27.9 / 28.0 \\

&
512
& 88.0
& 4.064
& 62.0
& 9.098
& 28.0 / 28.1 \\

&
1024 / 768
& 92.0
& 4.239
& 69.8
& 10.989
& 28.3 / 28.3 \\

\bottomrule
\end{tabular}%
}

\vspace{-3em}
\end{wraptable}

\newpage
\paragraph{Evaluation Benchmarks.} We evaluate diffusion stitching on a mix of mathematical reasoning and program synthesis tasks under a strictly zero-shot setting. For math, we report accuracy on GSM8K-CoT \cite{cobbe2021trainingverifiers}, \textsc{Math500} (a 500-problem subset of \textsc{MATH}~\cite{lightman2023lets}), and Countdown~\footnote{huggingface.co/datasets/Jiayi-Pan/Countdown-Tasks-3to4} (target-number arithmetic) (see Supplementary). For code, we report pass@1 on HumanEval~\cite{chen2021evaluating}, HumanEval+~\cite{evalplus}, MBPP~\cite{austin2021program}, and MBPP+ \cite{evalplus}. Unless stated otherwise, we use a fixed generation budget ($N{=}4$) with generation lengths in the range 256--512, and we follow each benchmark's standard answer normalization and scoring protocol (strict/exact match for math; unit-test execution for code). Finally, we evaluate all models in the zero-shot setting.

\subsection{Main Results}

\paragraph{Accuracy on reasoning benchmarks.}
\Cref{tab:tidar} summarizes overall task performance. 
In this table, we report Minerva Math to enable a direct comparison with TiDAR, while we use MATH500 for the detailed accuracy–latency and ablation studies later.
The results show that stitching yields large gains over vanilla diffusion decoding by converting diverse but noisy reasoning paths into a higher-quality evidence set: In the 8B setting, we outperform the LLaDA baseline decoded with a high confidence threshold ($\gamma = 0.9$), despite using a much lower threshold ($\gamma = 0.7$), by an average of \textbf{23.9\%} average points (72.38 vs. 48.51), with especially strong jumps on HumanEval (\textbf{+30.1}), HumanEval+ (\textbf{+26.2}), and MBPP (\textbf{+32.7}). We also outperform recent hybrid diffusion--AR systems, exceeding TiDAR (Trust Diff) by \textbf{+7.1}. Notably, our training-free pipeline is competitive with strong AR baselines: compared to Qwen3-8B, we are \textbf{+4.3} points higher on average, driven by sizable gains on GSM8K (\textbf{+9.7}) and consistent improvements on HumanEval (\textbf{+5.7}) and HumanEval+ (\textbf{+7.3}) (with MBPP+ roughly matched). These results highlight that step-level reuse can deliver state-of-the-art accuracy among diffusion-based approaches while remaining latency-friendly: exploration is cheap and parallel in diffusion, and the generative solver is invoked only once to produce the final answer.

\vspace{-12pt}
\paragraph{Accuracy vs. inference steps.}
To study efficiency,~\cref{fig:pareto} reports Pareto frontiers as accuracy versus inference steps, where we measure inference steps as the number of model forward evaluations required to answer a problem (this includes the forward passes of the diffusion model, PRM, and AR Solver). Across MBPP, GSM8K, and \textsc{Math500}, diffusion stitching achieves a consistently better trade-off than both diffusion-only baselines (which are efficient but less accurate) and AR baselines (which are accurate but require many sequential decoding steps), demonstrating an effective test-time scaling regime for low-latency reasoning. 
Moreover, we achieve much higher reasoning performance with far fewer steps than prior unified architectures (TiDAR), suggesting that a natural decoupling of diffusion-based exploration and AR-based verification is an important design choice for efficient reasoning.

\subsection{Ablation Study}

\paragraph{Effect of Aggregation Strategy and the Generative Solver.}
We ablate both how we aggregate diffusion traces and when a generative solver is needed in~\cref{tab:ablate_stitching}. Concretely, we compare three setups. 
\emph{Without a solver}, we either (i) decode a single trajectory (\textbf{Baseline}) or (ii) apply self-consistency by majority voting over final answers from multiple generations (\textbf{Majority Vote}; \cite{wang2023selfconsistencyimproveschainthought}). 
\emph{With a solver but without stitching}, we condition an AR model on either (iii) the concatenation of all sampled CoTs (\textbf{All CoT}) or (iv) the single highest-quality CoT (\textbf{Best CoT}), selected by the geometric mean of PRM step scores. 
\emph{With a solver and stitching}, we provide the solver with (v) our confidence-filtered stitched rationale (\textbf{Stitching}), (vi) an improved variant that keeps all above-threshold steps and also the best CoT as an anchor (\textbf{Stitching + Best CoT}). 

Overall, we observe a clear progression: diverse exploration provides a stronger set of candidate trajectories, adding a generative solver improves coherence via re-computation, and confidence-based stitching best leverages the full set of sampled traces. These results are consistent across both GSM8K and \textsc{Math500}, where we see that the full pipeline obtains an average of $14.7\%$ improvement over the strong LLaDA baseline decoded with a high confidence ($\gamma = 0.9$).

\begin{table}
\centering
\begin{minipage}{0.48\linewidth}
\centering
\scriptsize
\caption{\textbf{Diffusion-only generation vs. AR solver.} We compare recomputation using an autoregressive solver against a diffusion model solver with different sampling confidence levels.}
\vspace{0.5em}
\begin{tabular}{lcccc}
\toprule
 & \multicolumn{2}{c}{\textbf{GSM8K}} & \multicolumn{2}{c}{\textbf{MATH500}} \\
\cmidrule(r){2-3} \cmidrule(l){4-5}
\textbf{Setting} & Acc & \#steps & Acc & \#steps \\
\midrule

\rowcolor{gray!12} \multicolumn{5}{l}{\emph{Stitching (4 traces)}} \\
\myalign{l}{\;\;\;\footnotesize $\rotatebox[origin=c]{180}{$\Lsh$}$ AR} & 91.5 & 96.9 & 54.2 & 135.5 \\
\myalign{l}{\;\;\;\footnotesize $\rotatebox[origin=c]{180}{$\Lsh$}$ dLLM} $\gamma = 0.7$ & 90.1 & 106.3 & 50.2 & 132.6 \\
\myalign{l}{\;\;\;\footnotesize $\rotatebox[origin=c]{180}{$\Lsh$}$ dLLM} $\gamma = 0.9$ & 90.3 & 141.6 & 52.0 & 186.5 \\









\midrule
\rowcolor{gray!12} \multicolumn{5}{l}{\emph{Stitching (8 traces)}} \\
\myalign{l}{\;\;\;\footnotesize $\rotatebox[origin=c]{180}{$\Lsh$}$ AR} & 91.4 & 110.8 & 55.0 & 141.1 \\
\myalign{l}{\;\;\;\footnotesize $\rotatebox[origin=c]{180}{$\Lsh$}$ dLLM} $\gamma = 0.7$ & 91.4 & 115.8 & 52.6 & 138.6 \\
\myalign{l}{\;\;\;\footnotesize $\rotatebox[origin=c]{180}{$\Lsh$}$ dLLM} $\gamma = 0.9$ & 92.5 & 151.8 & 54.4 & 198.2 \\

\bottomrule
\end{tabular}
\label{tab:diffusion_solver}

\end{minipage}
\hfill
\begin{minipage}{0.48\linewidth}
\centering
\scriptsize
\caption{\textbf{Ablation of stitching strategies}. We compare selecting the best rollout, majority voting over final answers, using all steps, and our confidence-filtered step stitching, showing that selective stitching yields the best accuracy. Using $\gamma = 0.7$ and $\tau = 1.4$ and $\tau = 0.8$ for GSM8K and \textsc{Math500} respectively.}
\vspace{0.5em}
\label{tab:ablate_stitching}
\begin{tabular}{lccc}
\toprule
\textbf{Setting} & \textbf{GSM8K} & \textbf{MATH500} \\
\midrule
\rowcolor{gray!12} \multicolumn{3}{l}{\emph{Baseline}} \\
LLaDA & 78.8 & 37.6\\
\midrule
\rowcolor{gray!12} \multicolumn{3}{l}{\emph{without Solver or PRM}} \\
Majority vote~\cite{wang2023selfconsistencyimproveschainthought} & 85.1 & 42.0 \\
\midrule
\rowcolor{gray!12} \multicolumn{3}{l}{\emph{with Solver and PRM}} \\
All CoT  & 86.6 & 46.0 \\
Best CoT & 90.1 & 49.2 \\
\midrule
\rowcolor{gray!12} \multicolumn{3}{l}{\emph{with Solver, PRM, and Stitching}} \\
Above confidence & 90.4 & 52.0 \\
\rowcolor{cyan!15} Best CoT and above confidence & \textbf{91.5} & \textbf{54.2} \\
\bottomrule
\end{tabular}
\end{minipage}
\end{table}

\paragraph{Low-latency inference via parallel generations.}
Our framework supports low-latency test-time scaling because reasoning traces are independent and can be executed in parallel across devices. ~\cref{fig:teaser} plots accuracy versus end-to-end latency on \textsc{Math500} and shows that diffusion stitching achieves a better trade-off than both strong AR baselines, which achieve high accuracy but incur high sequential decoding latency, and vanilla diffusion baselines, which are fast but less accurate. By combining cheap parallel exploration with PRM-based step selection and a lightweight recomputation stage, stitching closes the gap to AR reasoning while remaining in the low-latency regime.

\begin{table}
\centering
\begin{minipage}{0.48\linewidth}
\small
\caption{\textbf{Ablation on the confidence threshold}. We report the average steps for the longest diffusion trajectory of each question + the number of solver steps required to regenerate the final answer. We use $\tau = 1.4$ and $\tau = 0.8$ for GSM8K and \textsc{Math500} respectively with the main experiments setup being highlighted in \colorbox{cyan!15}{blue}.}
\vspace{0.5em}
\label{tab:ablate_confidence}
\begin{tabular}{ccccc}
\toprule
 & \multicolumn{2}{c}{\textbf{GSM8K}} & \multicolumn{2}{c}{\textbf{MATH500}} \\
\cmidrule(lr){2-3}\cmidrule(lr){4-5}
\textbf{Confidence} & Acc & \#steps & Acc & \#steps \\
\midrule
0.4 & 87.6 & 76.6 & 46.8 & 103.4 \\
0.5 & 89.1 & 79.8 & 49.2 & 108.7 \\
0.6 & 90.5 & 86.8 & 52.0 & 119.8 \\
\rowcolor{cyan!15} 0.7 & 91.5 & 96.9 & 54.2 & 135.5 \\
0.8 & 91.7 & 111.3 & 54.4 & 158.5 \\
\bottomrule
\end{tabular}
\end{minipage}
\hfill
\begin{minipage}{0.48\linewidth}
\centering
\small
\caption{\textbf{Ablation on the number of generations}. Accuracy (\%) increases with more reasoning traces, and strong performance is already achieved with a modest number of samples. When these generations run in parallel, increasing the number of traces does not increase latency. The default number of traces used in the main experiments is highlighted in \colorbox{cyan!15}{blue}.}
\vspace{0.5em}
\label{tab:ablate_generations}
\begin{tabular}{lccc}
\toprule
& \textbf{GSM8K} & \textbf{MATH500} & \textbf{Avg} \\
\midrule
1  & 82.4 & 46.0 & 64.2 \\
2  & 88.1 & 49.2 & 68.7 \\
\rowcolor{cyan!15} 4  & 91.5 & 54.2 & 72.9 \\
6  & 91.4 & 54.3 & 72.9 \\
8  & 91.3 & 55.0 & 73.2 \\
\bottomrule
\end{tabular}
\end{minipage}
\end{table}

\paragraph{Deployment trade-offs.}
The latency advantage of parallel diffusion exploration comes with increased
memory usage. As shown in Table~\ref{tab:ar_latency}, our Qwen-PRM configuration
uses 42.8/42.7\,GB peak memory on GSM8K/MATH500, compared with approximately
14.4\,GB for the corresponding single-model AR baselines and 28\,GB for
PRM best-of-4. Replacing the verifier with the smaller Skywork-1.5B PRM
reduces peak memory to 32.3/32.2\,GB and also lowers latency, at the cost of
1.9 and 2.5 accuracy points on GSM8K and MATH500, respectively. Thus, our
method is most attractive when parallel hardware is available and latency is
more constrained than aggregate memory usage. The modular design also exposes
a direct accuracy--latency--memory trade-off through the choice of verifier
and number of parallel trajectories.

\paragraph{Scaling with the number of generations.}
We next study how performance scales with the number of diffusion generations $N$ on GSM8K and \textsc{Math500} (see~\cref{tab:ablate_generations}). Increasing $N$ improves performance initially, as additional trajectories provide more opportunities to recover correct intermediate steps and reduce the impact of any single noisy path. This trend is most pronounced on \textsc{Math500}, where each trace often shares re-usable sub-derivations but diverge in later steps; stitching allows us to preserve those shared correct components while avoiding late-stage derailment.

\paragraph{Diffusion sampling confidence.}
A key advantage of our framework is that it remains accurate even when the diffusion sampler is run with a \emph{low sampling confidence}, producing highly noisy and low quality trajectories. Our pipeline is explicitly designed to tolerate this noise: the PRM scores each step and filter out those of low quality. The stitched rationale can also contain gaps, yet the downstream generative solver can still recover the final solution.

\cref{tab:ablate_confidence} shows that accuracy is stable across a wide range of sampling confidences, while the required inference steps change substantially. On GSM8K, reducing the sampling confidence from $0.8$ to $0.6$ decreases the average step budget from $111.3$ to $86.8$ (a $22.0\%$ reduction) with little loss in accuracy (from $91.7$ to $90.5$). On \textsc{Math500}, the same reduction lowers steps from $158.5$ to $119.8$ (a $24.4\%$ reduction), again with comparable accuracy ($54.4$ vs.\ $52.0$). These results indicate that we can operate the diffusion model with a relatively low confidence to substantially reduce latency.

\vspace{-0.5em}
\paragraph{Diffusion-only generation.}
Although the solver introduces only modest additional inference overhead relative to the overall reasoning process, we do explore replacing the autoregressive (AR) solver with the same diffusion language model for evaluating whether diffusion can jointly support exploration and final answer synthesis.
The results are shown in table \ref{tab:diffusion_solver}, where we see that the diffusion solver benefit more from additional traces, reflecting their sensitivity to incomplete or inconsistent evidence. Sampling confidence also plays a different role: low-confidence sampling helps exploration but harms final answer quality, while high-confidence decoding is essential for stable recomputation. Overall, this highlights a clear asymmetry: diffusion models excel at exploration but struggle with consolidation; even with more trajectories and higher-confidence decoding, they still slightly underperform AR solvers, consistent with findings from TIDAR~\cite{Liu2025TiDAR}.

\paragraph{Other diffusion backbones.}
Our stitching framework is model-agnostic: it only assumes (i) a generator that can produce candidate reasoning trajectories (here, masked diffusion LMs) and (ii) a PRM that can score intermediate steps. To test generality, we apply our framework to several other diffusion backbones.
 
\begin{wraptable}{l}{0.58\textwidth}
\centering
\small
\captionof{table}{\textbf{dLLM/AR model ablation}. AR models are accurate but require many decoding steps, whereas our method improves dLLM performance while reducing inference cost by enabling lower-confidence sampling.}
\label{tab:ablate_models}
\begin{tabular}{lcccc}
\toprule
& \multicolumn{2}{c}{\textbf{GSM8K}} & \multicolumn{2}{c}{\textbf{MATH500}} \\
\cmidrule(lr){2-3}\cmidrule(lr){4-5}
\textbf{Model} & Acc & \#steps & Acc & \#steps \\
\midrule
Qwen-Math-Instruct & 94.3 & 440 & 79.6 & 1504\\
Phi-4 & 92.4 & 295 & 76.8 & 791 \\
\midrule
LLaDA-1.5 & 84.0 & 101 & 38.0 & 163 \\
\myalign{l}{\;\;\;\footnotesize $\rotatebox[origin=c]{180}{$\Lsh$}$ w/ stitching} & \textbf{90.8} & 87 & \textbf{53.8} & 134 \\
\midrule
Dream & 82.6 & 512 & 48.2 & 512 \\
\myalign{l}{\;\;\;\footnotesize $\rotatebox[origin=c]{180}{$\Lsh$}$ w/ stitching} & \textbf{85.4} & 521 & \textbf{53.2} & 523 \\
\midrule
%
LLaDA 2.0 & 88.0 & 256 & 44.6 & 256 \\
\myalign{l}{\;\;\;\footnotesize $\rotatebox[origin=c]{180}{$\Lsh$}$ w/ stitching} & \textbf{89.6} & 266 & \textbf{61.4} & 267 \\
\bottomrule
\end{tabular}
\end{wraptable}

\cref{tab:ablate_models} shows consistent gains across diffusion models. For \textsc{LLaDA-1.5}, stitching raises \textsc{Math500} accuracy from $38.0$ to $53.8$ while reducing inference steps from $163$ to $134$, since it allows a lower confidence threshold $\gamma$ to be used. For \textsc{Dream}, stitching similarly increases accuracy on both GSM8K ($82.6 \rightarrow 85.4$) and \textsc{Math500} ($48.2 \rightarrow 53.2$), with essentially unchanged step counts (the diffusion model is already run at a fixed-length budget in this setting). Finally, on \textsc{LLaDA-2.0}, stitching provides a substantial boost on \textsc{Math500} ($44.6 \rightarrow 61.4$) while keeping compute comparable ($256 \rightarrow 267$ steps), and preserves strong performance. 

\paragraph{PRM reliability and failure analysis.}
We further study how PRM errors and incomplete evidence affect the final solver on MATH500. Among 26 examples whose stitched evidence contained at least one incorrectly selected step, the solver still recovered the correct answer in 8 cases (30.8\%), showing that recomputation can tolerate some verifier errors. To identify the source of incorrect predictions, we intervene on 50 failures by adding a single missing intermediate step and rerunning the solver. As shown in Table~\ref{tab:prm_failure}, only 1/50 failures is attributable to a PRM false negative; 7/50 arise because the required step was never generated, while most cases (40/50) cannot be repaired by adding only one missing step. This suggests that incomplete diffusion exploration, rather than isolated PRM errors, is the dominant failure mode.

We also directly evaluate verifier quality on 2,016 candidate steps from 60 MATH500 examples, labelled as mathematically correct or incorrect using three blinded GPT-5.6 judge passes. Qwen-Math-PRM-7B achieves higher precision, recall, and F1 than Skywork-1.5B, consistent with its stronger downstream accuracy. Among steps selected exclusively by each verifier, 150/250 Skywork-only selections are incorrect compared with 79/331 Qwen-only selections. Thus, verifier quality affects the evidence presented to the solver, but most final-answer failures arise from missing or incomplete reasoning rather than individual PRM mistakes.

\begin{table}[H]
\centering
\small
\caption{\textbf{PRM reliability and failure analysis on \textsc{Math500}.}
Top: attribution of 50 incorrect predictions after adding one candidate missing reasoning step and rerunning the solver.
Bottom: step-level verifier quality on 2,016 candidate steps from 60 examples.}
\vspace{0.5em}
\label{tab:prm_failure}
\begin{tabular}{lccc}
\toprule
\rowcolor{gray!12} \multicolumn{4}{l}{\emph{Failure attribution}} \\
\textbf{Failure source} & \textbf{Count} & \textbf{Percent} & \\
\midrule
PRM false negative & 1 & 2.0\% & \\
Diffusion exploration failure & 7 & 14.0\% & \\
Stitching / solver failure & 2 & 4.0\% & \\
Not repaired by one missing step & 40 & 80.0\% & \\
\midrule
\rowcolor{gray!12} \multicolumn{4}{l}{\emph{Step-level verifier quality}} \\
\textbf{PRM} & \textbf{Precision} & \textbf{Recall} & \textbf{F1} \\
\midrule
Qwen-Math-PRM-7B & 88.97\% & 89.23\% & 89.10\% \\
Skywork-1.5B & 82.81\% & 78.17\% & 80.42\% \\
\bottomrule
\end{tabular}
\end{table}

\section*{Limitations}
Our framework improves reasoning efficiency primarily through parallel diffusion sampling, which increases memory usage and hardware requirements as the number of simultaneous trajectories grows. In addition, while stitching improves the accuracy–latency trade-off, the overall performance still depends on the quality of the underlying diffusion backbone and reward model used during inference.

\section{Conclusion}
We introduced a training-free reasoning framework that combines inexpensive diffusion-based exploration with step-level verification and selective aggregation. By scoring intermediate steps and stitching those with high confidence across independent rollouts, we turn noisy trajectories into a reusable pool of reliable evidence, which conditions a final generative solver to produce the answer. Across both maths and coding benchmarks, stitching substantially strengthens accuracy–latency trade-offs: it improves accuracy by up to +30.6\% over vanilla diffusion decoding and remains competitive with strong autoregressive baselines (e.g., +4.3\% on average), while achieving up to a 3.2× reduction in forward passes which is important for latency-critical settings. In summary, these results suggest that diffusion language models can support more efficient and scalable test-time reasoning by reusing high-quality intermediate computations across diverse reasoning trajectories, while also addressing domains where low latency and time-to-first-token are important constraints.
Overall, our results position diffusion stitching as a practical approach for latency-constrained reasoning when parallel inference is available. This latency advantage comes with increased memory and hardware requirements


\bibliography{references, references_more}
\bibliographystyle{abbrv}

\clearpage
\onecolumn
\appendix
\clearpage
\appendix
\onecolumn

\section{Supplementary Material}
This supplementary provides implementation and evaluation details that are omitted from the main paper due to space.
We first describe the models used to evaluate the quality of each reasoning step and to synthesize the final answers (Appendix~\ref{app:models_used}).
The next section (Appendix~\ref{app:temperature}) ablate the role of the temperature parameter in controlling the diversity of reasoning exploration.
After this, we then present the pseudo-code for the complete stitching procedure in algorithmic form (Appendix~\ref{app:pseudocode}), including the exact step-pool construction, confidence filtering, and evidence formatting provided to the generative solver.
We then report all hyperparameters and decoding settings needed to reproduce the reported results (Appendix~\ref{app:hyperparams}).
After this, we include additional comparisons against an RL fine-tuning baseline (Appendices \ref{app:rl}) and a recent speculative decoding baseline (Appendices \ref{app:specdec}).
Finally, we then list the prompts used for (i) diffusion generation, (ii) PRM step scoring, and (iii) solver recomputation (Appendix~\ref{app:prompts}).

\vspace{-0.5em}
\section{Models used.}
\label{app:models_used}

Unless stated otherwise, we use \textsc{LLaDA} as the masked diffusion backbone for generating candidate reasoning trajectories. We score intermediate steps with an off-the-shelf process reward model, and produce the final answer using an autoregressive solver conditioned on the stitched rationale. 

\noindent\textbf{Process reward models (PRMs).}
\begin{itemize}[leftmargin=1.2em, itemsep=0.1em, topsep=0.2em]
    \item \textsc{Qwen-Math-PRM-7B}~\cite{prmlessons}: step-level scoring for mathematical reasoning traces.
    \item \textsc{AceCoderRM}~\cite{zeng2025acecoderacingcoderrl}: reward model used to score each step in a block of code.
\end{itemize}

\noindent\textbf{Generative solvers.}
\begin{itemize}[leftmargin=1.2em, itemsep=0.2em, topsep=0.2em]
    \item \textsc{Qwen2.5-Math-Instruct}~\cite{yang2024qwen25mathtechnicalreportmathematical}: instruction tuned model for mathematical reasoning tasks.
    \item \textsc{Qwen2.5-Coder-7B}~\cite{hui2024qwen2}: code generation, code reasoning, and code fixing.
    \item \textsc{LLaDA 8B}~\cite{Nie2025LLaDA}: generalist masked diffusion language model.
\end{itemize}

\section{Controlling the diversity of reasoning paths.}
\label{app:temperature}
\begin{figure}[h]
    \centering
    \begin{subfigure}[b]{0.4\linewidth}
        \centering
        \includegraphics[width=\linewidth,clip,trim={0 0 265 0}]{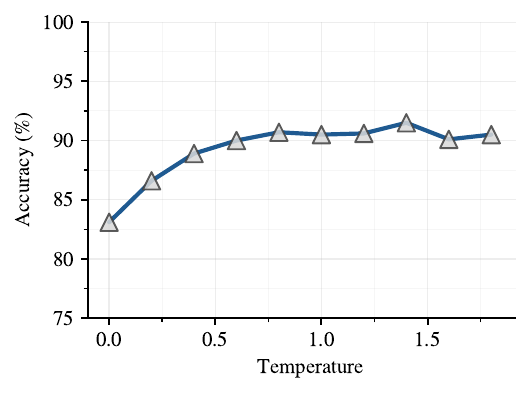}
        \caption{GSM8K}
    \end{subfigure}
    \hspace{0.05\linewidth} 
    \begin{subfigure}[b]{0.4\linewidth}
        \centering
        \includegraphics[width=\linewidth,clip,trim={265 0 0 0}]{figures/ablation_temp.pdf}
        \caption{\textsc{Math500}}
    \end{subfigure}
    \caption{\textbf{Ablation on sampling temperature}. Accuracy (\%) improves with higher temperature by promoting diverse reasoning paths. Although strong performance is achieved across a range of values, the optimal temperature is dataset-specific.}
    \label{fig:ablate_temperature}
\end{figure}

The benefits of stitching depend not only on how many reasoning traces are sampled but also on how diverse those traces are (see~\cref{fig:ablate_temperature}). If the trajectories are highly similar, additional samples add little new information. We control diversity using the temperature parameter for sampling, and evaluate the resulting trade-offs on GSM8K and \textsc{Math500}. We find that a moderate-to-high diversity is important for strong performance: too low temperatures leads to poor exploration and diversity, while too high degrades the quality of each trace.

\subsection{Segmentation, Anchor Choice, and Step Order}
\label{sec:stitching_heuristics}
The exact trace-scoring function has little effect on performance
(Table~\ref{tab:stitching_heuristics} top), suggesting that the geometric mean is not a
critical component of the method. In contrast, anchor quality is more
important: retaining the highest-scoring full rollout provides a complete
reasoning chain alongside high-confidence steps from the remaining rollouts
and consistently outperforms weaker, random, or absent anchors
(Table~\ref{tab:stitching_heuristics} middle). Finally, step segmentation and ordering have
a larger impact (Table~\ref{tab:stitching_heuristics} bottom); the canonical ordering
outperforms random or confidence-based sorting, while the alternative
segmentation substantially reduces performance, particularly on the
mathematical reasoning benchmarks.

\begin{table}[H]
\centering
\scriptsize
\caption{\textbf{Robustness of stitching design choices.}
We ablate the trace-scoring function, anchor selection, step ordering, and
segmentation strategy across math and coding benchmarks.}
\vspace{0.4em}
\label{tab:stitching_heuristics}
\begin{tabular}{lccc}
\toprule
\textbf{Setting} & \textbf{GSM8K} & \textbf{MATH500} & \textbf{MBPP} \\
\midrule

\rowcolor{gray!12}
\multicolumn{4}{l}{\emph{Step scoring}} \\
\rowcolor{cyan!15}
Geometric mean (ours) & 91.5 & 54.2 & \textbf{74.6} \\
Arithmetic mean       & \textbf{91.6} & \textbf{54.4} & 73.6 \\

\midrule
\rowcolor{gray!12}
\multicolumn{4}{l}{\emph{Anchor choice}} \\
\rowcolor{cyan!15}
Strong anchor (ours) & \textbf{91.5} & \textbf{54.2} & \textbf{74.6} \\
Weakest anchor       & 89.7 & 48.8 & 73.1 \\
Random anchor        & 90.2 & 50.4 & 72.6 \\
No anchor            & 90.4 & 52.0 & 73.6 \\

\midrule
\rowcolor{gray!12}
\multicolumn{4}{l}{\emph{Step ordering and segmentation}} \\
Random step order        & 90.2 & 53.0 & 73.0 \\
Confidence order         & 90.8 & 51.4 & 73.2 \\
\rowcolor{cyan!15}
Canonical ordering (ours) & \textbf{91.5} & \textbf{54.2} & \textbf{74.6} \\
Alternative segmentation  & 83.6 & 39.8 & 71.5 \\

\bottomrule
\end{tabular}
\end{table}

\subsection{Harder Reasoning Benchmark: AIME25}
\label{sec:aime}
We additionally evaluate on AIME25 using LLaDA2.0-mini (16B) to test the
pipeline on a harder reasoning benchmark. The diffusion backbone alone reaches
33.33\%, while a Qwen3 solver restricted to non-thinking mode reaches 23.33\%.
Combining parallel diffusion exploration with a short non-thinking Qwen3
recomputation improves accuracy to 46.67\%, showing that the solver can
effectively consolidate reasoning generated by the diffusion model without
performing a long reasoning trace itself. Qwen3 in thinking mode can achieve
higher accuracy with a larger token budget, but at substantially greater
latency.

\begin{table}[H]
\centering
\scriptsize
\caption{\textbf{Results on AIME25.}
For our method, the token budget is reported as diffusion generator / final
solver. The Qwen3 solver in our pipeline is used in non-thinking mode to
consolidate the stitched reasoning evidence and output the final answer.}
\vspace{0.4em}
\label{tab:aime25}
\setlength{\tabcolsep}{4pt}
\begin{tabular}{lcccc}
\toprule
\textbf{Method} & \textbf{Mode} & \textbf{Token limit} &
\textbf{AIME25} & \textbf{Latency (s)} \\
\midrule
LLaDA2.0-mini
    & -- & 8192 & 33.33 & 91.43 \\

Qwen3 solver-only
    & Non-thinking & 4096 & 23.33 & 40.35 \\

Qwen3 solver-only
    & Thinking & 8192 & 16.70 & 148.75 \\

Qwen3 solver-only
    & Thinking & 16392 & \textbf{56.70} & 233.13 \\

\rowcolor{cyan!15}
Ours (LLaDA2.0-mini + Qwen3)
    & Non-thinking & 8192 / 256 & 46.67 & 112.54 \\

\bottomrule
\end{tabular}
\end{table}

\newpage
\section{Full Stitching Algorithm}
\label{app:pseudocode}
Algorithm~\ref{alg:pseudocode} provides the pseudocode of our pipeline.
\renewcommand\thelstlisting{1}
\begin{figure}[h]
\centering
\begin{minipage}{0.9\linewidth}
\begin{lstlisting}[basicstyle=\scriptsize\ttfamily, mathescape, language=Python,
caption={PyTorch-style pseudocode for diffusion stitching},
captionpos=t, numberbychapter=false, label={alg:pseudocode}]
# p_theta : diffusion generator (exploration)
# PRM_phi : process reward model (step scoring)
# p_psi   : Generative solver (final recomputation)
# N       : num generations (traces)
# delta   : PRM stitch threshold (step selection)
# geom_mean(r) = (prod_t r[t])^(1/max(len(r),1))

for batch in loader:
  x, _ = batch                               # one problem x (batch size 1)
  S, R = [], []                              # steps and PRM scores per trace

  for n in range(N):                         # generations can be dispatched to different GPUs
    y_n  = diffusion_generate(p_theta, x)
    s_n  = extract_steps(y_n)                # s_n = [s^{(n)}_1, ..., s^{(n)}_{T_n}]
    r_n  = PRM_phi(x, s_n)                   # r_n = [r^{(n)}_1, ..., r^{(n)}_{T_n}] in [0,1]
    S.append(s_n); R.append(r_n)

  n_star = argmax_n geom_mean(R[n])          # best trace by geometric mean PRM score

  E = []                                     # stitched evidence list
  for n in range(N):
    for t in range(len(S[n])):
      keep = (R[n][t] >= delta) or (n == n_star)   # keep all best-trace steps as anchor
      if keep:
        E.append((t, R[n][t], S[n][t]))      # (step index, score, step text)

  E      = sort(E, key=(t, -score))          # chronological; break ties by higher score
  prompt = build_prompt(x, E)                # format as: [c=score] step
  y_hat  = ar_generate(p_psi, prompt)        # stop after complete \boxed{...}
\end{lstlisting}
\end{minipage}
\end{figure}

\section{Hyperparameters}
\label{app:hyperparams}
We report the exact models and score metrics used for each evaluation benchmark in table \ref{tab:hyperparams}.

\begin{table}[H]
\centering
\scriptsize
\setlength{\tabcolsep}{8pt}
\caption{\textbf{Metrics used for each evaluation task}. All tasks were evaluated zero-shot, using 4 generations, and with a generation length of 512. Strict match is implemented using the math-verify library\protect\footnote{github.com/huggingface/Math-Verify} and exact match uses string matching equality.}
\vspace{0.2em}
\renewcommand{\arraystretch}{1.15}
\begin{tabular}{l l l l l}
\toprule
\textbf{Task Category} & \textbf{Task Name} & \textbf{PRM} & \textbf{Solver} & \textbf{Score Metric} \\
\midrule
\multirow{4}{*}{Coding}
& HumanEval & \textsc{Qwen2.5-Math-PRM-7B} & \textsc{QWEN2.5-CODER-7B} & Pass@1 \\
& HumanEval+ & \textsc{Qwen2.5-Math-PRM-7B} & \textsc{QWEN2.5-CODER-7B} & Pass@1 \\
& MBPP & \textsc{ACECODERRM} & \textsc{QWEN2.5-CODER-7B} & Pass@1 \\
& MBPP+ & \textsc{ACECODERRM} & \textsc{QWEN2.5-CODER-7B} & Pass@1 \\
\midrule
\multirow{2}{*}{Math}
& GSM8K-CoT & \textsc{Qwen2.5-Math-PRM-7B} & \textsc{Qwen2.5-Math-Instruct} & strict-match \\
& MATH500 & \textsc{Qwen2.5-Math-PRM-7B} & \textsc{Qwen2.5-Math-Instruct} & strict-match \\
& Minerva Math & \textsc{Qwen2.5-Math-PRM-7B} & \textsc{Qwen2.5-Math-Instruct} & strict-match \\
\midrule
\multirow{1}{*}{Puzzle}
& Countdown & \textsc{Qwen2.5-Math-PRM-7B} & \textsc{Qwen2.5-Math-Instruct} & exact-match \\
\bottomrule
\end{tabular}
\label{tab:hyperparams}
\end{table}

\newpage
\section{More stitching examples}
\label{app:stitching_examples}
\begin{figure}[H]
    \centering
    \includegraphics[width=\linewidth]{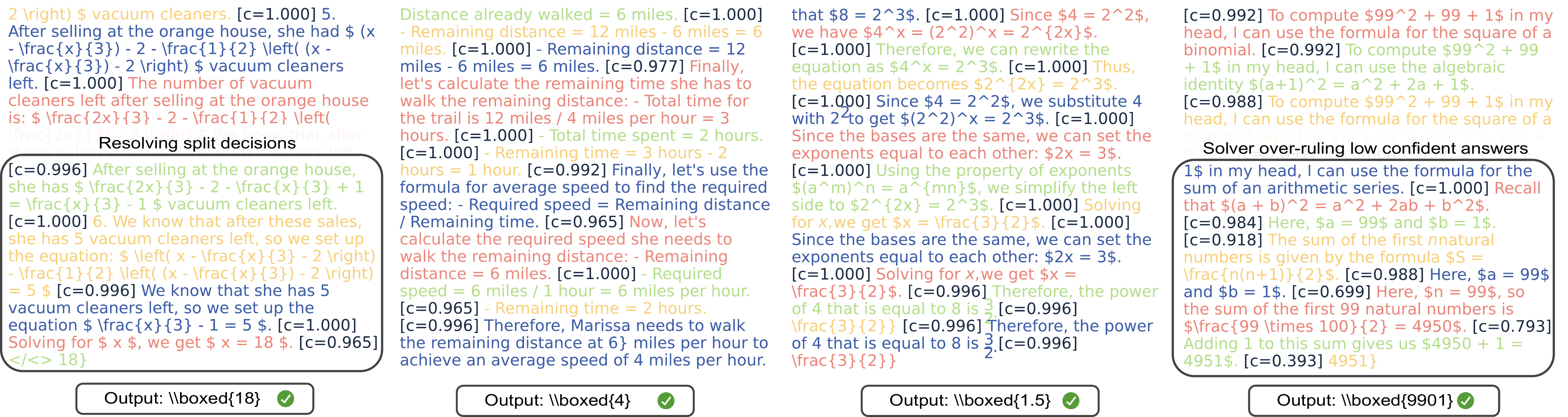}
    \caption{\textbf{Qualitative examples}. Left: resolving split decisions and right: over-ruling uniformly low-confidence predictions.}
    \label{fig:qualitative_success}
\end{figure}

Figure~\ref{fig:qualitative_success} shows representative cases where stitching succeeds. On the left, different diffusion trajectories contain complementary correct steps but disagree on the final answer; the solver resolves the split decisions into a correct final answer. On the far right, a less common case, the diffusion generations produce only low-confidence predictions, yet the solver over-rules them by recomputing from the stitched intermediate evidence.

\section{Failure cases}
\label{app:failure_cases}
\begin{figure}[H]
    \centering
    \includegraphics[width=\linewidth]{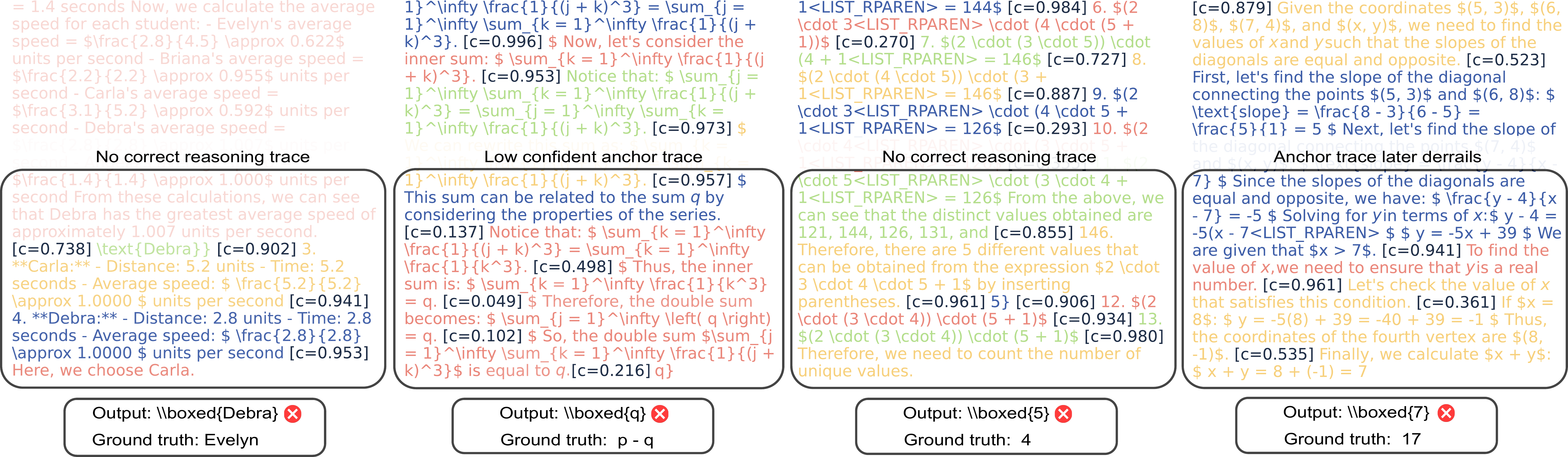}
    \caption{\textbf{Representative failure cases for stitching}. Each column shows the stitched evidence list and the final autoregressive solver prediction. Failures occur when the sampled trajectories contain no correct sub-derivation (left, third), when a key correct step is scored too low by the PRM and is omitted (second), or when the PRM is over-confident in an incorrect late step (or anchor trace), causing the evidence set to be dominated by a wrong conclusion (right).}
    \label{fig:qualitative_failures}
\end{figure}

To understand the limits of stitching, we include representative cases where it provides the incorrect final answer (Figure~\ref{fig:qualitative_failures}). These failures arise when the sampled diffusion trajectories lack a correct sub-derivation e.g., many traces share the same early error due to insufficient diversity, which results in the the stitched pool containing no reliable evidence to recompute a correct solution. Stitching can also fail when the PRM misranks steps, either assigning low scores to a crucial but correct intermediate step (so it is filtered out) or being over-confident in a fluent but incorrect late step, which then dominates the stitched evidence and steers the solver toward the wrong answer.

\newpage
\section{Comparisons with RL}
\label{app:rl}
We also compare against d1~\cite{zhao2025d1scalingreasoningdiffusion}, an RL fine-tuning approach that optimizes a model directly for performance on specific evaluation tasks. Unlike our training-free inference-time pipeline, d1 is a fine-tuning approach for specialist models using task specific reward losses. We compare in Table \ref{tab:d1}, where we can see that our generalist stitching framework is much more effective across various generation lengths, and evaluations.

\begin{table*}[ht]
\centering
\small
\caption{\textbf{Model performance on GSM8K, MATH500, and Countdown Benchmarks:}
All models are evaluated zero-shot with generation lengths of 256–512. Diffusion stitching is training-free, whereas d1 requires task-specific RL fine-tuning.}
\label{tab:d1}
\begin{tabular}{l*{6}{c}}
\toprule
 & \multicolumn{2}{c}{\textbf{GSM8K (0-shot)}} & \multicolumn{2}{c}{\textbf{MATH500 (0-shot)}} & \multicolumn{2}{c}{\textbf{Countdown (0-shot)}} \\
\cmidrule(lr){2-3}\cmidrule(lr){4-5}\cmidrule(lr){6-7}
\textbf{Model / Seq Len} & 256 & 512 & 256 & 512 & 256 & 512 \\
\midrule
d1-LLaDA
& 81.1 & 82.1
& 38.6 & 40.2
& 32.0 & 42.2 \\
\rowcolor{cyan!15} Ours
& \textbf{89.4} & \textbf{91.5}
& \textbf{49.4} & \textbf{54.2}
& \textbf{37.5} & \textbf{45.7} \\
\bottomrule
\end{tabular}
\end{table*}

\section{Comparisons with Speculative Decoding}
\label{app:specdec}
We compare against a recent speculative decoding baseline (RSD~\cite{liao2025rewardguidedspeculativedecodingefficient}), which follows a \emph{draft-and-verify} paradigm: a lightweight drafter proposes multi-tokens and a larger verifier selectively accepts them based on a scalar quality signal, falling back to the verifier to regenerate rejected segments. In contrast, diffusion stitching is not a token-acceptance scheme. We first use a diffusion model to \emph{explore} multiple complete reasoning traces in parallel, then \emph{select} and recombine high-quality intermediate steps into a stitched evidence list, and finally invoke an autoregressive solver \emph{once} to recompute the final answer conditioned on this evidence.

The results are shown in table \ref{tab:specdec}, where RSD attains only a small increase in accuracy on GSM8K, but requires substantially more forward passes. RSD uses 325 drafter steps and 75 verifier steps, whereas our pipeline uses 86 total steps (21.5\% of RSD).

\begin{table}[h]
\small
\centering
\caption{\textbf{Comparison to speculative decoding on \textsc{GMS8K.}} RSD steps are reported as \textit{drafter steps / verifier steps}. For ours, the steps are reported as \textit{diffusion steps / solver steps}.}
\vspace{0.5em}
\label{tab:specdec}
\begin{tabular}{lcc}
\toprule
& \multicolumn{2}{c}{\textbf{GSM8K}} \\
\cmidrule(lr){2-3}
\textbf{Model} & Acc & \#steps \\
\midrule
RSD~\cite{liao2025rewardguidedspeculativedecodingefficient} & 94.6 & 325/75 \\
\rowcolor{cyan!15} Ours & 91.5 & 80/6 \\
\bottomrule
\end{tabular}
\end{table}


\section{Impact Statement}
This paper presents work whose goal is to advance the field of machine learning. There are many potential societal consequences of our work, none of which we feel must be specifically highlighted here.

\newpage
\section{Prompts}
\label{app:prompts}
We use three prompt templates for the math reasoning benchmarks:

\begin{promptbox}{Maths: Diffusion Proposer Prompt (Generating Reasoning Traces)}
\normalsize
\textbf{System:} 
\begin{verbatim}
You are a math expert. You will be given a question to solve. 
Wrap the final answer in a \boxed{}.
Respond in the following format:
<reasoning>
Your reasoning here
</reasoning>
<answer>
\boxed{...}
</answer>
\end{verbatim}
\textbf{User:} \{PROBLEM\}
\end{promptbox}

\begin{promptbox}{Maths: PRM Scoring Prompt (Step-Level)}
\normalsize
\textbf{System:} 
\begin{verbatim}
Please reason step by step, and put your final answer within \\boxed{}.
\end{verbatim} \\
\textbf{User:} Problem: \{PROBLEM\}\\[4pt]
\textbf{Assistant:} \{ANSWER\} with \texttt{<extra\_0>} inserted after each step.
\end{promptbox}

\begin{promptbox}{Maths: Generative Solver Prompt (Condition on Stitched Rationale)}
\normalsize
\textbf{System:} 
\begin{verbatim}
Solve the problem. Use the candidate steps as evidence. 
If steps conflict, choose a consistent subset. 
Output only the final answer in \\boxed{...}.
\end{verbatim} \\
\textbf{User:} Problem: \{PROBLEM\}\\[4pt]
\textbf{Assistant:}
\begin{verbatim}
Here are candidate reasoning steps from multiple traces.
Each item: [c=conf]. Use them as evidence; ignore contradictions; 
prefer higher conf.
{STITCHED ANSWER}
</reasoning>
<answer>
\end{verbatim}
\end{promptbox}

\newpage
We use the following prompt templates for the coding benchmarks:

\begin{promptbox}{Code (MBPP, MBPP+): Diffusion Proposer Prompt (Generating Reasoning Traces)}
\normalsize
\textbf{System:} 
\begin{verbatim}
You are a helpful Python coding assistant. 
When given a programming problem, respond with self-contained 
Python 3 code that solves the task and passes the provided tests. 
Output only a single ```python ...``` code block with no additional 
commentary.

Problem:
{problem}

Details:
{description}

Unit tests your solution must pass:
{tests}

Setup code that runs before the tests:
{setup}

Write the full solution now. 
Remember to return only Python code inside a single fenced block.
\end{verbatim}
\textbf{User:} \{PROBLEM\}
\end{promptbox}

\begin{promptbox}{Code (HumanEval, HumanEval+): Diffusion Proposer Prompt (Generating Reasoning Traces)}
\normalsize
\textbf{System:} 
\begin{verbatim}
You are helping a separate autoregressive code model solve
HumanEval-style Python tasks.

Given the exact task prompt below (signature + docstring + examples), 
produce a short, concrete implementation guide.

REQUIREMENTS:
- Do NOT write any Python code.
- Do NOT include Markdown fences or explanations outside the hint block.
- Output 6-12 lines total.
- Each line must start with "# " (hash + space), with no other
  leading text.
- Focus on: algorithm choice, key steps, edge cases, complexity, and any 
  Python gotchas.
- Prefer actionable steps over prose. Do not restate the entire prompt.

TASK PROMPT:
\end{verbatim}
\textbf{User:} \{PROBLEM\}
\end{promptbox}

\newpage
\begin{promptbox}{Code: Generative Solver Prompt (Condition on Stitched Rationale)}
\normalsize
\textbf{System:} 
\begin{verbatim}
You are an expert Python programmer. 
You will be given a Python function signature and docstring. 
Write a correct implementation. 
Use the candidate steps as evidence. 
If steps conflict, choose a consistent subset.
\end{verbatim} \\
\textbf{User:} Problem: \{PROBLEM\}\\[4pt]
\textbf{Assistant:}
\begin{verbatim}
Here are candidate reasoning steps from multiple traces.
Each item: [c=conf]. Use them as evidence; ignore contradictions; 
prefer higher conf.
{STITCHED ANSWER}
\end{verbatim}
\end{promptbox}


\end{document}